# Efficient and Accurate MRI Super-Resolution using a Generative Adversarial Network and 3D Multi-Level Densely Connected Network

Yuhua Chen[1,2], Feng Shi[2], Anthony G. Christodoulou[2], Yibin Xie[2], Zhengwei Zhou[2] and Debiao Li[1,2]

[1] University of California, Los Angeles, CA 90095, USA
[2] Biomedical Imaging Research Institute, Cedars-Sinai Medical Center, CA 90048, USA
`chyuhua@ucla.edu,simonsf@gmail.com, {Anthony.Christodoulou,`
`Yibin.Xie, Zhengwei.Zhou, Debiao.Li}@cshs.org`

**Abstract.** High-resolution (HR) magnetic resonance images (MRI) provide detailed anatomical information important for clinical application and quantitative image analysis. However, HR MRI conventionally comes at the cost of longer scan time, smaller spatial coverage, and lower signal-to-noise ratio (SNR). Recent studies have shown that single image super-resolution (SISR), a technique to recover HR details from one single low-resolution (LR) input image, could provide high quality image details with the help of advanced deep convolutional neural networks (CNN). However, deep neural networks consume memory heavily and run slowly, especially in 3D settings. In this paper, we propose a novel 3D neural network design, namely a multi-level densely connected super-resolution network (mDCSRN) with generative adversarial network (GAN)–guided training. The mDCSRN trains and inferences quickly, and the GAN promotes realistic output hardly distinguishable from original HR images. Our results from experiments on a dataset with 1,113 subjects shows that our new architecture outperforms other popular deep learning methods in recovering 4x resolution-downgraded images and runs 6x faster.

**Keywords:** MRI, Super-Resolution, Image Enhancement, Deep Learning.

## 1    Introduction

High spatial resolution MRI produces detailed structural information, benefiting clinical diagnosis, decision making, and accurate quantitative image analysis. However, due to hardware and physics limitations, high-resolution (HR) imaging comes at the cost of long scan time, small spatial coverage, and low signal to noise [1]. The ability to restore an HR image from a single low-resolution (LR) input would potentially overcome these drawbacks. Therefore, single image super-resolution (SISR) is an attractive approach, as it requires only a LR scan to provide an HR output without extra scan time. But SR is a challenging problem because of its underdetermined nature [2]. An infinite number of HR images can produce the same LR image after resolution degradation. This makes it very difficult to accurately restore texture and structural details. A large portion of previous methods frame SR as a convex optimization problem, to find a plausible HR solution while balancing regularization terms [1, 3]. However, regularization terms require a priori knowledge of the image distribution, often based on experimental assumptions. Popular constraints like total variation implicitly assume that the image is piecewise constant, which is problematic for images with many local details and tiny structures. On the other hand, learning-based approaches do not require such well-defined priors. Especially, deep learning-based techniques have shown great improvement in SISR for images with abundant details, because of its non-linearity and extraordinary ability to imitate accurate transformation between LR and HR in difficult cases. Super-Resolution Convolutional Neural Networks (SRCNNs) [4] and their more recent Faster-SRCNNs (FSRCNNs) [5] draw a lot of attention as they showed that simple structured CNNs can produce outstanding SISR results of 2D natural images.

However, those previous adapted deep-learning approaches do not fully solve the puzzle in the medical image SR problem. First, many medical images are 3D volumes, but previous CNNs only work slice by slice, discarding information from continuous structures in the third dimension. Second, 3D models have far more parameters than 2D models, raising a challenge in memory consumption and computational expenses, making them less practical. Finally, the most widely used optimization objective of CNN is pixel/voxel-wise error like mean squared error (MSE) between model estimation and the reference HR. But as mentioned in [6], MSE and its derivative Peak Signal to Noise Ratio (PSNR) do not directly represent the visual quality of restored images. Thus, using MSE as the only target leads to overall blurring and low perceptual quality.

In this paper, we propose a 3D Multi-Level Densely Connected Super-Resolution Networks (mDCSRN) to fully solve the above problems. By utilizing a densely connected network [7], our mDCSRN is extremely light-weight. When optimized by intensity difference, it provides the state-of-art performance while keeping the model much smaller and faster. Then when trained with a Generative Adversarial Network (GAN), it improves further, outputting sharper and more realistic-looking images.

## 2 Method

Our proposed SISR neural network model aims to learn the image prior for inversely mapping the LR image to the reference HR image. The model only takes LR images to produce SR images. During the training, HR reference will be used to guide the optimization of the model's parameters. During deployment, SR images can be generated by the model based on the input LR. Details are provided in the followings:

### 2.1 Background

The resolution downgrading process from an HR image $X$ to a LR image $Y$ can be presented as:

$$Y = f(X), \tag{1}$$

where $f$ is the function causing a loss of resolution. The SISR process is to find an inverse mapping function $g(\cdot) \approx f^{-1}(\cdot)$ to recover HR image $\hat{X}$ from a LR image $Y$:

$$\hat{X} = g(Y) = f^{-1}(Y) + R, \tag{2}$$

where $f^{-1}$ is the inverse of $f$ and $R$ is the reconstruction residual.

In a CNN SISR approach, three different steps are optimized together: feature extraction, manifold learning, and image reconstruction. During the training, the difference between reconstructed images and ground truth images is not only used to adjust reconstruction layer to restore better images from manifold, but also to guide extraction of accurate image features. This mingling of different components makes it possible for neural network to achieve state-of-art performance among other SISR techniques [4].

### 2.2 Training a SR network with GAN

The most intuitive way to optimize the reconstruction is by minimizing the voxel wise difference such as absolute difference (L1 loss) or mean square error (L2 loss). However, minimizing L1 or L2 loss leads to solutions which resemble a voxel-wise average of possible HR candidates, which does not penalize the formation of artificial image features at the neighbor or patch level. Thus, the output tends to be over-blurred and implausible to the human eye. For better optimization, we incorporated the idea from Ledig et al [6] to use a Generative Adversarial Network (GAN)–based loss function.

The GAN framework has two networks: a generator $G$ and a discriminator $D$. The basic idea of a GAN is to train a $G$ to produce images with rich details while simultaneously training a $D$ to distinguish the given image as either real or generated. At the end of the training, $D$ will be a very good classifier to separate real and generated images, while the $G$ can generate realistic looking images according to $D$. The advantage of using GAN is that it can be optimized without a predesigned loss function for a specific task. In SISR, SRGAN was proposed by [6], who showed that adding GAN's $D$ loss to guide the $G$'s training yields high perceptual quality.

However, training of a GAN presents its own challenges. During training, $G$ and $D$ must be balanced to evolve together. If either of them becomes too strong, the training will fail, and $G$ can learn nothing from $D$. For 2D natural images, a lot of effort have been made to stabilize the training process. However, these approaches greatly rely on the network structure and have yet to be described for newer architectures like Dense-Net. To stabilize the training process, the Wasserstein GAN (WGAN) authors [8] observed that the failure of GAN training is due to its optimization toward Kullback-Leibler divergence between real and generated probability. When there is little or no overlap between them, which is very common in the early stage of training, the gradient from the discriminator will vanish and the training will stall. To address this issue, WGAN was proposed. Its loss function approximately optimizes Earth Mover (EM) distance, which can always guide the generator forward. WGAN enables almost fail-free training and produces quality as good as vanilla GAN. Additionally, the EM distance between real and generated images from $D$ can be regarded an indicator of the image quality. In this work, we used WGAN for additional guiding during training.

### 2.3 Need for Efficient 3D Super-Resolution Network

It has been shown that 3D super-resolution models outperforming 2D counterparts by a large margin, thanks to the fact that 3D model directly learns the 3D structure of MRI volumetric images [9]. However, one significant drawback of a 3D model is that 3D deep learning model usually has a much larger number of parameters due to the extra dimensions of convolutional filters. For example, a relatively shallow 2D FSRCNN has only 12,000 parameters while its 3D version has 65,000, a >5x difference. The number of parameters determines the model size and computation cost of a deep learning network, which is a key issue to consider for practical use.

Recently, DenseNet has shown that by using dense skip connections, we can dramatically reduce the network size while maintaining state-of-art performance in natural image classification. Yet, even memory-efficient DenseNets have too many parameters when constructed in 3D. The basic idea of densely connection from DenseNet was applied here, but we also include a new architecture that uses an extra level of skip connections. This not only helps to reduce the parameter number but also speeds up the computation. We discuss the detailed design of mDCSRN in the following section.

### 2.4 Proposed 3D Multi-Level Super-Resolution Network

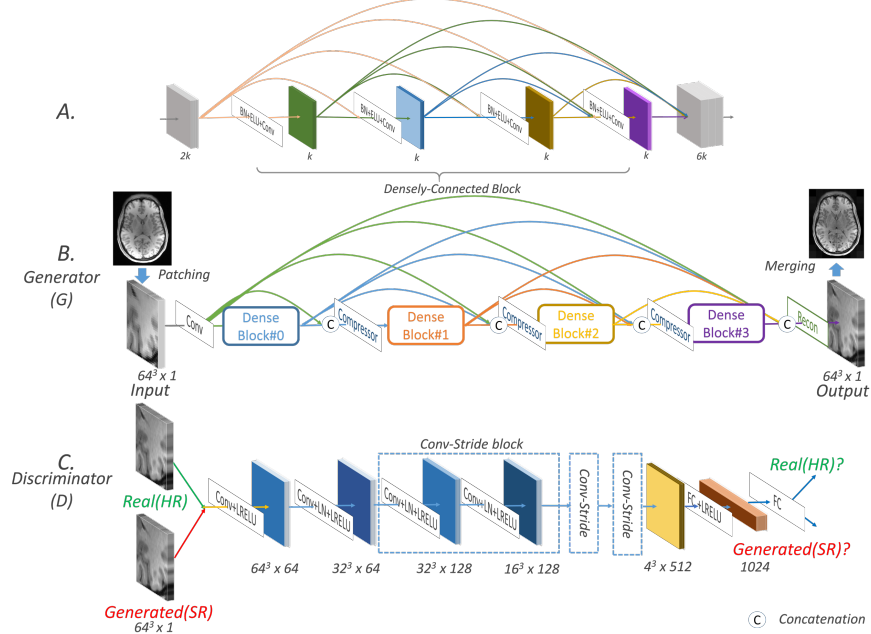

**Fig. 1.** Architecture of (A) DenseBlock with 3x3x3 convolutions and (B, C) mDCSRN-GAN Network. The *G* is *b4u4*(4 blocks, each has 4 unites) mDCSRN. The first convolutional layer outputs 2k (k=16) feature maps, and each compressor shrinks down the feature maps to 2k via a 1x1x1 convolution. The final reconstruction layer is an another 1x1x1 convolution. The *D* is identical to SRGAN except BatchNorm is replaced by LayerNorm, suggested by WGAN-GP.

A recent study [9] shows that Densely Connected Super-Resolution Network (DCSRN) with a single DenseBlock, is already capable of capturing image features and restoring super-resolution images, outperforming other state-of-art techniques. But further improvement of the network performance is required to make use of a deeper model to catch more complex information in SR process. However, the memory consumption of a DenseNet increases dramatically as the number of layers increases, which makes it not feasible to train or deploy a deeper DCSRN.

To address this problem, we propose a multi-level densely connected structure, where a single deep DenseBlock is split into several shallow blocks. As shown in Fig.1(B), each DenseBlock takes the output from all previous DenseBlock and is directly connected to reconstruction layer, following the same principle of DenseNet. Those skip connections provide direct access to all former layers including the input enables uninterrupted gradient flow, which is proven to be more efficient and less overfit. However, unlike original DenseNet, there is no pooling layer in mDCSRN, so mDCSRN can make full use of information in full resolution.

Another improvement is to add a 1x1x1 convolutional layer as a compressor before all the following DenseBlocks. One key attribute to empower deep learning models to generalize so well is that the model has an information compression that forces the model to learn universal features to avoid overfitting. In our design, the compressors bottleneck the network to the same width for each DenseBlock. This is expected to provide at least two benefits: 1) To reduce the memory consumption from hyperbolically to linearly dependent on depth; 2) To equally weight each DenseBlock, preventing later DenseBlocks (which takes care of conceptual level image features) from dominating the network with more parameters, thereby forcing the network not to overlook local image features that are central to the super-resolution task.

### 2.5 Design of Loss Function

In our work, we utilized gradient penalty variants of WGAN, namely WGAN-GP to speed up the training convergence. Our loss function has two parts: intensity loss $loss_{\text{int}}$ and GAN's discriminator loss $loss_{\text{GAN}}$:

$$loss = loss_{\text{int}} + \lambda loss_{\text{GAN}} \tag{3}$$

where $\lambda$ is a hyperparameter that we set to 0.001. We used the absolute difference (L1 loss) between network output SR and ground truth HR images as the intensity loss:

$$loss_{\text{int}} = loss_{\text{L1}}/\text{LHW} = \sum_{z=1}^{L}\sum_{y=1}^{H}\sum_{x=1}^{W}\left|I_{x,y,z}^{\text{HR}} - I_{x,y,z}^{\text{SR}}\right|/LHW \tag{4}$$

where $I_{x,y,z}^{\text{SR}}$ is the super-resolution output from the deep learning model and $I_{x,y,z}^{\text{HR}}$ is the ground truth HR image patch. We use GAN's discriminator loss as the additional loss to the SR network:

$$loss_{\text{GAN}} = loss_{\text{WGAN},D} = -D_{\text{WGAN},\theta}(I^{\text{SR}}) \tag{5}$$

where $D_{\text{WGAN},\theta}$ is the discriminator's output digit from WGAN-GP for SR images.

### 2.6 LR Images Generation

To evaluate an SR approach, we need to generate LR images from ground truth HR images. LR images are generated following the same steps as in [9]: 1) converting HR image into k-space by applying FFT; 2) downgrading the resolution by truncating outer part of 3D k-space with a factor of 2x2; 3) converting back to image space by applying inverse FFT and linearly interpolating to the original image size. This mimics the actual acquisition of LR and HR images by MRI scanners.

## 3 Experiments

### 3.1 Dataset and Data Preparation

To better demonstrate the generalization of the deep learning model, we used a large publicly accessible brain structural MRI database, the human connectome project. 3D T1W images from a total of 1,113 subjects were acquired via Siemens 3T platform using 32-channel head coil on multiple centers. The images come in high spatial resolution as 0.7 mm isotropic in a matrix size of 320x320x256. These high-quality ground truth images provide detailed small structures, which is a perfect case for SR project. The whole dataset is split into 780 training, 111 validation, 111 evaluation and 111 test samples by subject. No subjects nor image patches appear twice in different subsets. Validation set is used for monitoring training process and evaluation set is used for hyper-parameters selection. We only use test set for final performance evaluation to avoid fine-tuning model favorable to test set data.

The original images were used as ground-truth HR images, and then degraded to LR. We used the exact same process of patching and data augmentation as in [9]. However, we merged the patches without overlapping, which makes model run even faster and results in less blurring. We left a margin of 3 pixels to avoid artifacts on the edge.

### 3.2    Training Parameters and Experiment Setting

The models were implemented in Tensorflow on a workstation with Nvidia GTX 1080TI GPU. The DenseBlock in mDCSRN setting is similar with DCSRN, where all 3D convolutional layers had filter with size 3x3x3, growth rate k=16. For comparison, we picked up relatively small network FSRCNN [9] and more complicated state-of-art SRResNet [6]. We selected the same hyper-parameters according to 2D FSRCNN [5]. And we extended the 2D convolution to 3D for both FSRCNN and SRResNet.

For non-GAN networks, ADAM optimizer with a learning rate $10^{-4}$ was used to minimize the L1 loss function with a batch size of 2. We trained for 500k steps as no significant improvement afterward. For GAN experiments, we transfer the weights from well-trained mDCSRN in non-GAN training as the initial G. For the first 10k steps, we trained the discriminator only. After then for every 7 steps of training discriminator, we trained the generator once; and every 500 steps we train discriminator for an extra 200 step alone, which makes sure that discriminator is always well-trained, as suggested in WGAN. Adam optimizer with $5\times10^{-6}$ is used to optimize both G network for 550k steps, with little improvement after that.

To demonstrate the effectiveness of mDCSRN compared with DCSRN, we made four different network setups with varied block number(b) and unit number(u). A network with single 8-unit DenseBlock is annotated as *b1u8* and a network with four DenseBlocks each has 4 dense-units is annotated as *b4u4*, respectively. We used three image metrics: subject-wise average structural similarity index (SSIM), peak signal to noise ratio (PSNR), and normalized root mean squared error (NRMSE), to measure the similarity between SR image and reference HR image in the 2x2 down-sampled plane.

### 3.3    Results

The quantitative results from non-GAN approaches are shown in **Table 1**. The parameters and running speed of each networks are also listed in Table 1. DCSRN *b1u8* and mDCSRN *b2u4* had the same depth of network, but the later obtained marginally better results and reduce parameters and running time by more than 30%. Among all variants, the largest network *b4u4* has the best performance without too much sacrifice in speed.

**Table 1.** The results of SSIM, PSNR and NRMSE for different DCSRN architectures. With the same depth, *b2u4* has a slightly better performance than *b1u8* with less number of parameters and computation operation. The deepest network *b4u4* had an average runtime for a whole 3D MRI of a subject just around 20 seconds while has the best performance.

| | DCSRN b1u8 | | | mDCSRN b2u4 | | | mDCSRN b3u4 | | | mDCSRN b4u4 | | |
|---|---|---|---|---|---|---|---|---|---|---|---|---|
| | SSIM | PSNR | NRMSE | SSIM | PSNR | NRMSE | SSIM | PSNR | NRMSE | SSIM | PSNR | NRMSE |
| *mean* | 0.9371 | 35.35 | 0.0906 | *0.9381* | *35.46* | *0.0895* | 0.9402 | 35.56 | 0.0884 | **0.9424** | **35.88** | **0.0852** |
| *std* | 0.0053 | 0.79 | 0.0038 | 0.0053 | 0.78 | 0.0038 | 0.0052 | 0.79 | 0.0038 | 0.0051 | 0.78 | 0.0038 |
| *#parm* | 0.307M | | | 0.200M | | | 0.304M | | | 0.412M | | |
| *#ops* | 1.247M | | | 0.813M | | | 1.236M | | | 1.672M | | |
| *Time(s)* | 13.20 | | | **9.74** | | | 15.13 | | | *20.87* | | |

**Table 2.** Performance comparison between bicubic interpolation, 3D FSRCNN, 3D SRResNet and our proposed mDCSRN *b4u4*. mDCSRN provides similar image quality to SRRestNet but 6x faster and provides much better image quality than bicubic interpolation and FSRCNN.

*FSRCNN has large CNN kernels (size: 5 and 9) that are extremely computationally expensive, though small #ops, it takes longer time than mDCSRN which only has small filters (size: 3).

| | Bicubic Interpolation | | | 3D FSRCNN* | | | 3D SRResNet | | | mDCSRN b4u4 | | |
|---|---|---|---|---|---|---|---|---|---|---|---|---|
| | SSIM | PSNR | NRMSE | SSIM | PSNR | NRMSE | SSIM | PSNR | NRMSE | SSIM | PSNR | NRMSE |
| *mean* | 0.8377 | 29.07 | 0.1873 | 0.9211 | 34.11 | 0.1045 | *0.9412* | *35.71* | *0.0869* | 0.9424 | 35.88 | 0.0852 |
| *std* | 0.0088 | 0.90 | 0.0087 | 0.0059 | 0.77 | 0.0042 | 0.0052 | 0.79 | 0.0038 | 0.0051 | 0.78 | 0.0038 |
| *#parm* | - | | | 0.064M | | | 2.005M | | | 0.412M | | |
| *#ops* | - | | | 0.261M* | | | 8.043M | | | 1.672M | | |
| *Time(s)* | - | | | 21.27 | | | *132.71* | | | **20.87** | | |

mDCSRN *b4u4* was compared with bicubic interpolation as well as other neural networks FSRCNN and SRResNet (**Table 2.**). mDCSRN obtained a large advantage against FSRCNN methods and is slightly better than SRResNet but runs more than 6x faster. Additionally, our mDCSRN-GAN provides much sharpened and visually plausible images compared with non-GAN approaches. **Fig. 2** demonstrates super-resolution results of one random subject in the 2x2 resolution degrading plane. Among non-GAN methods, the small vessels in mDCSRN are more distinguishable than in other neural networks. However, mDCSRN-GAN provides much better overall image quality: not only does the vessel maintains the same shape and size as in the ground-truth HR image, but the gaps between vessel and gray matter are also much clearer (see red arrows). The mDCSRN-GAN result is almost indistinguishable from the ground truth.

## 4    Conclusion

We have presented a novel SISR method based on 3D mDCSRN-GAN for MRI. We showed that mDCSRN-GAN can recover local image textures and details more accurately, and 6 times more quickly than current state-of-art deep learning approaches. This new technique would allow 4-fold reduction in scan time while maintaining virtually identical image resolution and quality.

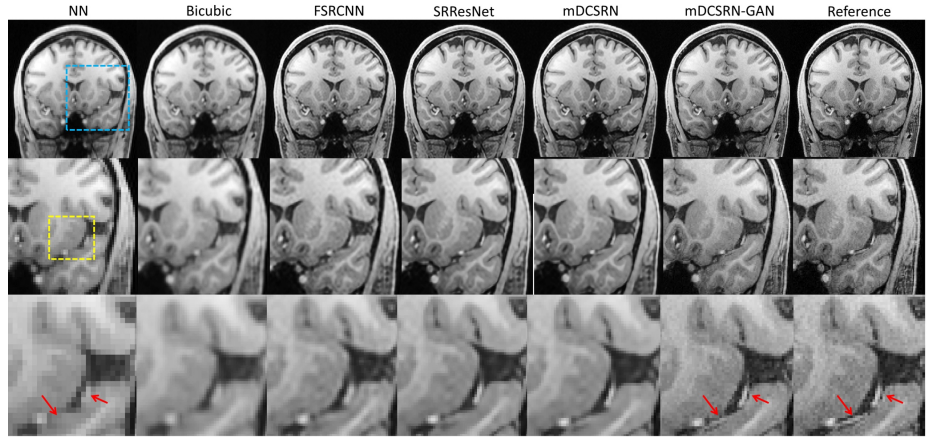

**Fig. 2.** Illustration of the nearest neighbor(NN) and bicubic interpolation, 3D FSRCNN, 3D SRResNet, mDCSRN, mDCSRN-GAN reconstruction results, and corresponding HR images.

## References

1. E. Plenge, D. H. J. Poot, M. Bernsen, G. Kotek, G. Houston, P.Wielopolski, L. V. D. Weerd, W. J. Niessen, and E. Meijering,"Super-resolution methods in MRI: Can they improve the trade-off between resolution, signal-to-noise ratio, and acquisition time?" Magnetic Resonance in Medicine, vol. 68, no. 6, pp. 1983–1993, Jan. 2012

2. Tanno R. et al. (2017) Bayesian Image Quality Transfer with CNNs: Exploring Uncertainty in dMRI Super-Resolution. In: Descoteaux M., Maier-Hein L., Franz A., Jannin P., Collins D., Duchesne S. (eds) Medical Image Computing and Computer Assisted Intervention − MICCAI 2017.  Lecture Notes in Computer Science, vol 10433. Springer, Cham

3. Shi, Feng, et al. "LRTV: MR image super-resolution with low-rank and total variation regularizations." IEEE transactions on medical imaging 34.12 (2015): 2459-2466.

4. C. Dong, C.C. Loy, K. He, and X. Tang, "Image Super-Resolution Using Deep Convolutional Networks," IEEE Transactions on Pattern Analysis and Machine Intelligence, vol. 38, no. 2, pp. 295–307, Jan. 2016

5. Dong, Chao, C.C. Loy, and Xiaoou Tang. "Accelerating the super-resolution convolutional neural network." European Conference on Computer Vision. 2016.

6. Ledig, Christian, et al. "Photo-Realistic Single Image Super-Resolution Using a Generative Adversarial Network." Proceedings of the IEEE Conference on Computer Vision and Pattern Recognition. 2017.

7. Huang, Gao, et al. "Densely connected convolutional networks." Proceedings of the IEEE conference on computer vision and pattern recognition. Vol. 1. No. 2. 2017.

8. Arjovsky, Martin, Soumith Chintala, and Léon Bottou. "Wasserstein gan." arXiv preprint arXiv:1701.07875 (2017).

9. Chen, Yuhua, et al, "Brain MRI super resolution using 3D deep densely connected neural networks," Proceedings of 15th International Symposium on Biomedical Imaging (ISBI 2018), pp. 739-742.